# A Decision-Based View of Causality

**David Heckerman**
Microsoft Research, Bldg 9S
Redmond, WA 98052-6399
heckerma@microsoft.com

**Ross Shachter**
Department of Engineering-Economic Systems
Stanford, CA 94305-4025
shachter@camis.stanford.edu

## Abstract

Most traditional models of uncertainty have focused on the associational relationship among variables as captured by conditional dependence. In order to successfully manage intelligent systems for decision making, however, we must be able to predict the effects of actions. In this paper, we attempt to unite two branches of research that address such predictions: causal modeling and decision analysis. First, we provide a definition of causal dependence in decision-analytic terms, which we derive from consequences of causal dependence cited in the literature. Using this definition, we show how causal dependence can be represented within an influence diagram. In particular, we identify two inadequacies of an ordinary influence diagram as a representation for cause. We introduce a special class of influence diagrams, called causal influence diagrams, which corrects one of these problems, and identify situations where the other inadequacy can be eliminated. In addition, we describe the relationships between Howard Canonical Form and existing graphical representations of cause.

## 1 Introduction

Most traditional models of uncertainty, including Markov networks [Lauritzen, 1982] and belief networks [Pearl, 1988] have focused on the associational relationship among variables as captured by conditional independence and dependence. Associational knowledge, however, is not sufficient when we want to make decisions under uncertainty. For example, although we know that smoking and lung cancer are probabilistically dependent, we cannot conclude from this knowledge that we will increase our chances of getting lung cancer if we start smoking. In general, to make rational decisions, we need to be able to predict the effects of our actions.

Recent work by Artificial Intelligence researchers, statisticians, and philosophers—for example, Pearl and Verma (1991), Druzdzel and Simon (1993), and Spirtes et al. (1993)—have emphasized the importance of identifying causal relationships for purposes of modeling the effects of intervention. They argue, for example, that if we believe that smoking causes lung cancer, then we know that if we start to smoke then we will increase our chances of getting lung cancer. In contrast, if we believe that there is a gene that causes both lung cancer and our desire to smoke, then we know that if we start to smoke then we will not increase our chances of getting lung cancer.

For over a decade, decision analysts have used the influence diagram to represent decisions problems [Howard and Matheson, 1981]. In doing so, they have worried about representing the effects of interventions (decisions) on a set of uncertain variables. Nonetheless, they have avoided using notions of causality in their work, in large part because no precise definition of causality has been proposed.

In this paper, we attempt to weave together these two threads of research. In particular, we propose a definition of causal dependence in clear decision-analytic terms, which we derive from consequences of such dependence often cited in the causal modeling literature. We thereby offer a means by which the results in each discipline may be translated to the other. Thus, for example, decision analysts may translate Pearl's (1994) calculus of intervention to a method for proving stochastic dominance. Conversely, researchers working on causal modeling can use influence diagrams in Howard Canonical Form for planning under uncertainty.

Given the audience of this paper, we concentrate mostly on translating results from decision analysis to the representation and manipulation of causal dependence. After defining causal dependence, we show how it can be represented within an influence diagram. We identify two inadequacies of the influence diagram as a representation for cause, and introduce a special class of influence diagrams, called causal influence diagrams, which corrects one of these problems. In addition, we identify situations where the other inadequacy can be eliminated. Also, we show how influence diagrams in Howard Canonical Form may be used to answer counterfactual queries. Finally, we show the relationship between Howard Canonical Form and Pearl and Verma's (1991) causal theory.

## 2 Background

Fundamental to our discussion is the distinction between a *chance variable* and *a decision variable*. In general, a variable has a (possibly infinite) set of mutually exclusive and collectively exhaustive possible *states*. The state of a decision variable is chosen by a person, usually called the decision maker. In contrast, a chance variable is uncertain and its state may be at most indirectly affected by the decision maker's choices. For example, the decision to smoke or not is a decision variable, whereas whether or not a person develops lung cancer is a chance variable. We shall use lowercase letters to denote single variables, and uppercase letters to denote sets of variables. We call an assignment of state to every variable in set $X$ an *instance* of $X$. Typically, we refer to the possible states of a decision variable as *alternatives*. We use a probability distribution $P\{X|Y\}$ to represent a decision maker's uncertainty about $X$, given that a set of chance and/or decision variables $Y$ is known or determined.

In this paper, we are interested in modeling relationships in a *domain* consisting of chance variables $U$ and decision variables $D$. We use the influence diagram to model these relationships. An *influence diagram* is (1) a directed acyclic graph containing decision and chance nodes corresponding to decision and chance variables, and information and relevance arcs, representing what is known at the time of a decision and probabilistic dependence, respectively, (2) a set of probability distributions associated with each node, and optionally (3) a set of utilities for all possible instances of $U \cup D$. A *belief network* is an influence diagram containing only chance nodes and relevance arcs.

An *information arc* is one that points to a decision node. An *information arc* from chance or decision node $a$ to decision node $d$ encodes the assertion that variable $a$ will be known when decision $d$ is made. (We shall use the same notation to refer to a variable and its corresponding node in the diagram.)

A *relevance arc* is one that points to a chance node. The *absence* of a relevance arc represents conditional independence. To identify relevance arcs, we ask the decision maker to order the variables in $U = (x_1, \ldots, x_n)$. Then, for each variable $x_i$, we ask the decision maker to identify a set $Pa(x_i) \subseteq \{x_1, \ldots, x_{i-1}, D\}$ that renders $x_i$ and $\{x_1, \ldots, x_{i-1}, D\} \setminus Pa(x_i)$ conditionally independent. That is,

$$P\{x_i|x_1, \ldots, x_{i-1}, D\} = P\{x_i|Pa(x_i)\} \quad (1)$$

For every variable $a$ in $Pa(x_i)$, we place a relevance arc from $a$ to $x_i$ in the diagram. That is, the nodes $Pa(x_i)$ are the *parents* of $x_i$. All conditional independencies implied by the given assertions can be read from the graph using d-separation [Pearl, 1988].

Associated with each chance node $x_i$ in an influence diagram are the probability distributions $P\{x_i|Pa(x_i)\}$. It follows that any influence diagram for $U \cup D$ uniquely determines a joint probability distribution for $U$ given $D$. A *deterministic node* is a special kind of chance node that is a deterministic function of its parents. A *minimal influence diagram* is an influence diagram where Equation 1 would be violated if any arc were removed.

Finally, an influence diagram may contain a single distinguished node, called a *utility node* that encodes the decision maker's utility for each instance of the node's parents.

Figure 1 contains an influence diagram for two lifestyle decisions: whether or not to smoke and whether or not to change diet. As is illustrated in the figure, we use ovals, squares, and a diamond to represent chance, decision, and utility nodes, respectively. There are no information arcs in the diagram, although we can imagine one between *smoke* and *diet* indicating the order in which the decisions are made.[1] The influence diagram contains several missing relevance arcs. One assertion made by the absence of these arcs is that *lung cancer* and *cardiovascular status* are conditionally independent, given *smoke*, *diet*, and *genotype*. This assumption and others in the diagram are questionable, but they will serve for purposes of example.

When modeling decision problems, as decision analysts well know, it is important to clearly define all variables. A good negative example of this is our description of variables in Figure 1. For example, what does *cardiovascular status* really mean? Does the variable refer to a patient's mean blood pressure over a period of a week, or to the amount of exercise that induces chest pain? Does the variable refer to this status one year from now or two? In order to facilitate the precise definition of variables, decisions analysts often use a gedanken experiment known as the *clarity test* [Howard, 1994, Chapter 7]. A variable is said to pass a *clarity test* if a clairvoyant, who can see the past, present, and future with certainty, can tell us which state the variable is in.[2] After applying the clarity test, the decision maker may define *cardiovascular status*—for example—to be the average diameter of her major coronary arteries as determined by angiography one year from the present.

---

[1] Decision order is not important for our discussion.

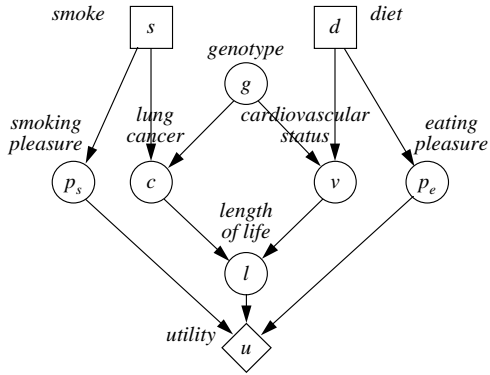

Figure 1: An influence diagram for decisions about lifestyles.

## 3 Fixed Sets and Cause

In this paper, we take a practical view for our definition of causality. Rather than ask "what is causality?," we ask "what do we use the notion of causality for?" Answering the first question is extremely difficult. Philosophers have been arguing over the answer for centuries. Answering the second question is straightforward. When a chance variable $x$ causes a chance variable $y$, we know that any intervention that results in the same state of $x$ will lead to the same state of $y$. Knowing this correspondence is useful for modeling decision problems. If we know that $x$ causes $y$, then we know that if we represent the relationship between $x$ and $y$, this relationship will remain unchanged in the face of decisions that change only $x$ directly. In this paper, we take this useful property of causality to be its definition.

To define causality, we must first formalize the more primitive notion of a fixed set.

**Definition 1 (Fixed Set)** Given
uncertain variables $U$ and decisions $D$, the fixed set for $U$ with respect to $D$, denoted $F(D)$, is the set of variables in $U$ that take on the same states regardless of the choice of $D$. The conditional fixed set for $U$ given $C$ with respect to $D$, denoted $F(D|C)$, is the set of variables in $U$ which take on the same instance for any given instance of $C$, regardless of the choice of $D$.

In our lifestyles example, it is reasonable to assert that *genotype* is in the fixed set for $D = \{smoke, \ diet\}$, because we expect that genotype will be the same regardless of the decisions made. Also, we may assert that *length of life* is in the conditional fixed set $F(D|cardiovascular\ status, \ lung\ cancer)$, because we believe that *length of life* will be the same, once we know *cardiovascular status* and *lung cancer*, regardless of the decisions made. Note that, in this example and in general, the order in which decisions $D$ are made is irrelevant to the identification of fixed sets.

Conditioning on $C$ in the definition of conditional fixed set is different from probabilistic conditioning. In particular, when we assess $P\{x|C\}$, we imagine that $C$ has been observed. When we assess $F(D|C)$, however, $C$ may depend on $D$; and so $C$ cannot be observed prior to our choice for $D$. Instead, we imagine that we have made decisions $D$ and *subsequently* observe $C$. We then identify the set of variables that take on the same instance for any possible choice of $D$ whenever $C$ takes on the same instance.

Membership in a conditional fixed set is closely related to conditional independence. Most important, both concepts are subjective. The assertion that $x$ is in the fixed set $F(D|C)$ belongs to a particular decision maker, just as does the assertion that $x$ and $D$ and conditionally independent given $C$. In addition, when $x$ and $D$ are conditionally independent given $C$, then one's *belief* in $x$ does not change with changes in $D$ if $C$ is known. When $x$ is in the conditional fixed set $F(D|C)$, then $x$ *itself* does not change with changes in $D$ once $C$ is known. Thus, fixed set membership is stronger than is conditional independence: If $x \in F(D|C)$, then $x$ and $D$ are conditionally independent given $C$. We note that $X \subseteq F(D|C)$ interpreted as the $\mathcal{I}$-statement $I(D,C,X)$ satisfies the graphoid axioms [Pearl, 1988] of decomposition, weak union, and contraction, but not symmetry.

If a chance variable $x$ is not in the fixed set with respect to $D$, then—to some extent—it is under the control of the decision maker. Consequently, neither the decision maker nor the clairvoyant can observe $x$ prior to the decisions $D$ being made. Conversely, if $x$ is in the fixed set with respect to $D$, then it is not under the control of the decision maker and may be observed—at least in principle. We call this observation the fundamental property of fixed set observation.

---

[2]The clairvoyant can only know the future conditioned on the decisions to be made. We address this point in the following section.

**Proposition 1 (Fixed Set Observation)** *Prior to making a commitment to a set of decisions D, it is impossible to observe those variables outside the fixed set $F(D)$, and we may observe those variables in that fixed set.*

We can now formalize our definition of cause.

**Definition 2 (Cause)** *Given uncertain variables U and decisions D, the variables C are causes for $x \in U$ with respect to D if C is a minimal subset of $D \cup U \setminus \{x\}$ such that (1) $x \notin F(D)$, and (2) $x \in F(D|C)$.*

The second condition formalizes the notion that $x$ does not change with a set of decisions $D$, given a set of chance variables $C$. The first condition guarantees that the second condition is not satisfied trivially. That is, the first condition guarantees that the decisions $D$ can affect $x$. The minimality condition prevents some causes of $C$ from being superfluous. Note that, by our definition, causes may be decisions and/or chance variables, but only chance variables may be caused.

Variants of our lifestyles decision problem shown in Figure 2 help us to illustrate the definition. Some conclusions that can be drawn about each domain are shown next to the influence diagram for that domain. First, let us consider the decision problem in Figure 2a. Here, we model only the decision of whether or not to smoke; and we do not bother to model the variable *length of life*. For this problem, it is reasonable to assert that *lung cancer* is not in the fixed set $F(smoke)$. Also, it is true trivially that *lung cancer* is in the conditional fixed set $F(smoke|smoke)$. Consequently, by our definition, we can conclude that $\{smoke\}$ is a singleton cause for *lung cancer*. Similarly, we may conclude that $\{smoke\}$ is a cause for *smoking pleasure*. In addition, it is reasonable to assert that *utility* is not in $F(smoke)$, *utility* is in the conditional fixed set $F(smoke|smoking\ pleasure,\ lung\ cancer)$, and there is no subset $C$ of $\{smoking\ pleasure,\ lung\ cancer\}$ such that *utility* is in $F(smoke|C)$. Therefore, we can conclude that $\{smoking\ pleasure,\ lung\ cancer\}$ are causes for *utility*. As shown in the figure, we may also conclude that $\{smoke\}$ is a cause for *utility*. This example illustrates an important property of our definition: causes are not unique.

Someday, it may be possible to use retroviral therapy to alter one's genetic makeup. Assuming that a decision of whether or not to undergo such therapy is available, it is reasonable to assert that *genotype* is not in the fixed set with respect to $D = \{smoke,\ retroviral\ therapy\}$. In contrast, it is reasonable to assert that *genotype* is in the conditional fixed set $F(D|retroviral\ therapy)$. Therefore, $\{retroviral\ therapy\}$ is a singleton cause for *genotype*. Furthermore, it is reasonable to assert that *lung cancer* is not in $F(D)$, that *lung cancer* is in $F(D|smoke,\ genotype)$, and that for no subset $C$ of $\{smoke,\ genotype\}$ is *lung cancer* in $F(D|C)$. Thus, we can conclude that $\{smoke,\ genotype\}$ are causes for *lung cancer*. This example demonstrates that the conclusions drawn about cause and effect, given our definition, depend strongly on what decisions are available. Thus, as in our formal definition, we say that $\{smoke,\ genotype\}$ are causes for *lung cancer* with respect to $\{smoke,\ retroviral\ therapy\}$.[3]

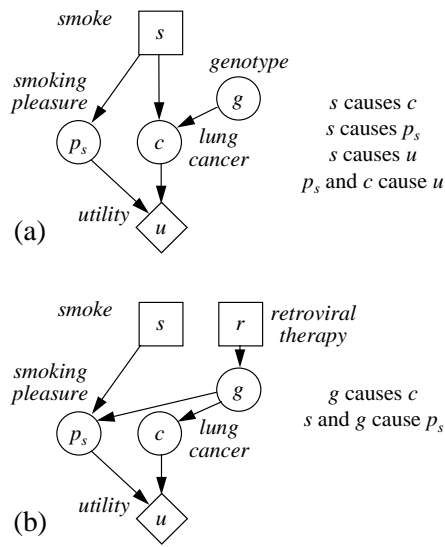

Figure 2: Variants the decision problem in Figure 1.

The influence diagram in Figure 2b corresponds to an "alternative" view of the relationships between the decision to smoke and lung cancer. Here, smoking is not a cause for lung cancer. Rather, genotype is a cause for lung cancer, and both genotype and smoking are causes for smoking pleasure. As mentioned in the introduction, these two views predict differently what would happen should one start smoking. This example emphasizes that our definition of cause and effect is subjective. One person may hold beliefs corresponding to the model in Figure 2a, whereas another person may hold beliefs corresponding to the model in Figure 2b. Both people are "correct" provided they make their decisions in a manner that is consistent with their beliefs.

Our definition of causality is satisfying for several reasons. One, it is consistent with a universally accepted notion of causality: an effect cannot precede its cause. In particular, we cannot observe a chance variable that is caused by a decision before we make that decision.

---
[3]To be brief, we often omit this last clause in our assertions of causal dependence.

Also, it satisfies the reasonable property that $x$ and $y$ cannot cause each other, except for the special case where $x$ and $y$ are related deterministically. Namely, if $\{x\}$ is a cause for $y$ with respect to $D$ and $\{y\}$ is a cause for $x$ with respect to $D$, then one can show that $x$ must be a deterministic function of $y$ and $D$ (and $y$ must be a deterministic function of $x$ and $D$).

We emphasize that our definition allows us to use causality to model actions more than it explains the fundamental notion of causality. We do not consider this a practical weakness, because unless we can intervene, recognition of causality has no benefit. For example, if a variable is in the fixed set with respect to our decisions, then there is no use in identifying what we would otherwise perceive as its causes.

## 4 Graphical Representation of Cause: Causal Influence Diagrams

Given the known benefits of the belief network for representing conditional independence, we should expect that a graphical representation of cause and effect would be useful. In the previous section, we saw that our notion of cause and effect is intimately related to the notion of the (conditional) fixed set. Thus, we desire a graphical representation in which we can encode the existence or lack thereof of fixed sets.

At first glance, the influence diagram appears to be such a representation. In particular, consider the following graphical condition.[4]

**Definition 3 (Block)** *Given an influence diagram with decision nodes $D$ and chance nodes $U$, $C \subseteq U$ is said to* block $D$ *from* $x \in U$ *if every directed path from a node in $D$ to $x$ contains at least one node in $C$.*

If we reexamine our examples in Figure 2, we see that whenever $x$ is not blocked from $D$ by the empty set—that is, whenever there is a path from a decision node to $x$—then $x$ is not in the fixed set $F(D)$. In addition, whenever $x$ is blocked from $D$ by $C$, then $x$ is in the fixed set $F(D|C)$. Thus, in these examples, we may read cause and effect directly from the influence diagram.

In other examples, however, this correspondence between the graphical condition of blocking and fixed sets breaks down. Consider the simple decision problem shown in Figure 3a. Here, we have the decision $d$ of whether or not to bet *heads* or *tails* on the outcome of a coin flip $c$. Whether or not we win is represented by the variable $w$. Note that $w$ is a deterministic function of $d$ and $c$ (indicated by the double oval). Suppose we believe that the coin is fair—that is p(heads)=1/2. In this case, if we do not bother to model the variable $c$ explicitly, as shown in Figure 3b, we need not place an arc from $d$ to $w$, because the probability of winning will be $1/2$, regardless of our choice $d$. Nonetheless, $w$ is not in the fixed set $F(d)$, because $w$ will take on different states for different bets. Consequently, we have a situation where there is no path from $d$ to $w$, and yet $w$ is not in the fixed set $F(d)$.

---

[4]As is made clear in the proofs of Theorems 1 and 2, we do not need the full d-separation criterion.

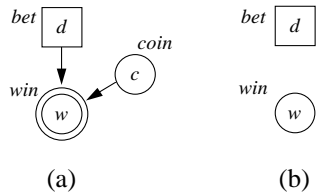

Figure 3: Influence diagrams for betting on a coin flip.

Conversely, consider a subset of our lifestyles decision problem shown in Figure 4a. If we choose not to model the variable *genotype*, we can obtain the influence diagram shown in Figure 4b. In this influence diagram, we cannot remove any arc without producing invalid assertions of conditional independence. Nonetheless, *cardiovascular status* is in the fixed set $F(D|diet)$.

In order to discuss clearly the inadequacies of the influence diagram for the representation of fixed sets, we introduce the following concepts, which closely parallel Pearl's concepts of $\mathcal{I}$-map and $\mathcal{D}$-map.

**Definition 4 ($\mathcal{F}$-map)** *An influence diagram is said to be an $\mathcal{F}$-map if*

$$C \text{ blocks } D \text{ from } x \implies x \in F(D|C)$$

**Definition 5 ($\overline{\mathcal{F}}$-map)** *An influence diagram is said to be an $\overline{\mathcal{F}}$-map if*

$$C \text{ does not block } D \text{ from } x \implies x \notin F(D|C)$$

In our coin example, the influence diagram in Figure 3b is an $\overline{\mathcal{F}}$-map, but not an $\mathcal{F}$-map. Conversely, in our smoking example, the influence diagram in Figure 4b is an $\mathcal{F}$-map, but not an $\overline{\mathcal{F}}$-map.

We can guarantee that an influence diagram is an $\mathcal{F}$-map by adding additional arcs to it. In the coin example, having learned from the decision maker that $w$ is not in the fixed set with respect to $d$, we can add an arc from $d$ to $w$, making the diagram an $\mathcal{F}$-map. In general, however, we do not want to burden a decision maker with the task of guaranteeing that all fixed set assertions implied by the blocking condition are valid. Fortunately, we can obtain such a guarantee, by asking the decision maker to provide a set of local assertions about fixed set membership. We call influence diagrams with these added local assertions *causal influence diagrams*.

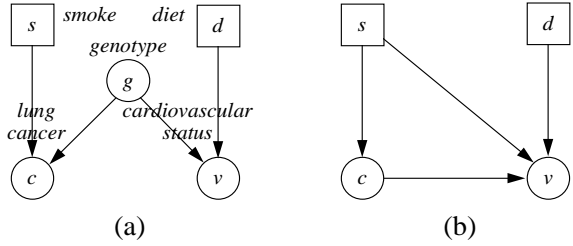

Figure 4: A subset of the lifestyles decision problem.

**Definition 6 (Causal Influence Diagram)** *A causal influence diagram is an influence diagram in which every node $x$ is in $F(D|Pa(x))$.*

**Theorem 1** *All causal influence diagrams are $\mathcal{F}$-maps.*

**Proof:** Suppose that a set of chance nodes $C$ block $D$ from $x$, but that $x$ is not in the conditional fixed set $F(D|C)$. Because the influence diagram is causal, it follows that at least one of $x$'s parents—say—$y$ would not be in $F(D|C)$. Applying this argument recursively, until $y \in C$, we obtain a contradiction. $\square$

The situation is not as simple for $\overline{\mathcal{F}}$-maps. Let us consider the following theorem.

**Theorem 2** *If an influence diagram is a $\mathcal{D}$-map, then it is an $\overline{\mathcal{F}}$-map.*

**Proof:** If $x \in F(D|C)$, then $x$ and $D$ are conditionally independent given $C$. Because the influence diagram is a $\mathcal{D}$-map, it follows that $C$ blocks $D$ from $x$ provided no head-to-head nodes or descendants of head-to-head nodes that are downstream from decisions are observed. If one such node—say—$y$ were observed, however, it would have to be in the fixed set $F(D)$. Therefore, $y$ and $D$ would be independent, contradicting the fact that there is a directed path from $D$ to $y$ and the influence diagram is a $\mathcal{D}$-map. $\square$

Given Theorems 1 and 2, we have the following sufficient conditions for representing cause and effect in an influence diagram.

**Theorem 3** *Given a causal influence diagram that is a $\mathcal{D}$-map, $C$ are causes for $x$ with respect to decisions $D$ if (1) $x$ is a descendant of some decision node in $D$, and (2) $C$ (which does not include $x$) is a minimal set that blocks $D$ from $x$.*

**Proof:** Because the influence diagram is a $\mathcal{D}$-map and hence an $\overline{\mathcal{F}}$-map, condition 1 implies that $x$ is not in the fixed set $F(D)$. Because the influence diagram is causal and hence an $\mathcal{F}$-map, condition 2 implies that $x$ is in the fixed set $F(D|C)$. $\square$

It is reasonable to expect that decision makers will be able to construct causal influence diagrams. Such construction merely requires that the decision maker provide a set of local assessments about fixed set membership. Unfortunately, an influence diagram may not be a $\mathcal{D}$-map. The decision problem in Figure 4b is one such example. There is a special case, however, where we can read causal relationships directly from an influence diagram. Following Verma and Pearl (1991), let us consider a special type of decision, called a set decision.

**Definition 7 (Set Decision)** *Given an influence diagram for uncertain variables $U$ and decisions $D$, a set decision for $x \in U$ with respect to $D$ is any decision node $s_x \in D$ such that (1) $s_x$ has alternatives "set $x$ to $k$" for each state $k$ of $x$ and "do nothing," and (2) $x$ is the only child of $s_x$.*

Given a set decision $s_x$, we can literally set $x$ to any of its states, or we can do nothing. When we set $x$ to one if its states, none of the other ancestors of $x$ contribute to the determination of $x$.

**Theorem 4** *Given a minimal causal influence diagram for uncertain variables $U$ and decisions $D$, and a variable $x \in U$ which has nonempty parents $Pa(x)$, if $D$ includes a set decision for each chance-node parent of $x$, then $Pa(x)$ are causes for $x$.*

**Proof:** Consider any node $x$. If $x$ has no chance-node parents, then $x$ is caused by its decision parents. If $x$ has chance-node parents, then because the influence diagram is minimal and there exist set decisions for each such parent of $x$, $x$ must not be in the fixed set with respect to $D$. Furthermore, because the influence diagram is minimal and causal, $Pa(x)$, which includes the decisions pointing to $x$, must be a minimal set $C$ such that $x$ is in the conditional fixed set $F(D|C)$. Consequently, $x$ is caused by its parents. $\square$

## 5 Howard Canonical Form and Causal Mechanisms

Before making important decisions, decision analysts investigate how useful it is to gather additional information. This investigation is typically done by computing the value of information about one or more chance nodes in the domain. To compute the *value of information* of observing a chance variable $x$ with respect to a decision $d$, one computes the decision maker's expected value given that $x$ is observed before the decision $d$ is made, and subtracts it from the decision maker's expected value given that $x$ is not observed before the decision is made. If the actual value of learning something about $x$ is less than the value of information about $x$, we know that it is not worth while to gather such information.

According to the fundamental property of fixed set observation (Proposition 1), it is not possible to observe a variable outside the fixed set with respect to $D$. From Theorem 2, this restriction translates to the restriction that no chance variable downstream from a decision may be observed before a decision is made, assuming that the influence diagram is a $\mathcal{D}$-map. This rule can be stated in the following terms: If an influence diagram is a $\mathcal{D}$-map, then it cannot contain directed cycles. In practice, this well-known rule is applied to all influence diagrams [Howard and Matheson, 1981].

Thus, in an influence diagram that is a $\mathcal{D}$-map, we cannot compute the value of information for any variable $x$ that is a descendant of a decision node. Fortunately, with additional assessments, we can transform a given influence diagram into one where we can compute the value of information for any chance node. Such a transformed influence diagram is said to be in Howard Canonical Form [Howard, 1994, Chapter 7].

**Definition 8 (Howard Canonical Form)** *An influence diagram is said to be in* Howard Canonical Form *(HCF) if (1) it is a causal influence diagram, and (2) every chance node that is a descendant of a decision node is a deterministic node.*

For example, consider the simple influence diagram in Figure 5a. The corresponding influence diagram in HCF is shown in Figure 5b. In this new influence diagram, we have added the node $c(s)$, which is a variable that represents all possible deterministic mappings between *smoke* and *lung cancer*—that is, the variable represents *lung cancer* as a function of *smoke*. The possible states of this variable are shown in Table 1. Also, by definition of $c(s)$, *lung cancer* becomes a deterministic function of *smoke* and $c(s)$. For example, if *smoke* is *yes* and $c(s)$ is in state 1, then *lung cancer* will be *yes*. The uncertainty in the relationship between *smoke* and *lung cancer*, formerly represented in the node *lung cancer*, now resides in the node $c(s)$. In effect, we have *extracted* the uncertainty in the causal relationship between these two variables, and moved this uncertainty to the node $c(s)$.

Several points about the transformation to HCF are worthy of mention. First, condition 1 is important, because it guarantees that the nodes we create are in the fixed set with respect to decisions. If these new nodes were not in the fixed set, then we could not compute their value of information. Although Howard does not use our language, he includes this condition in his definition [Howard, 1994, Chapter 7].

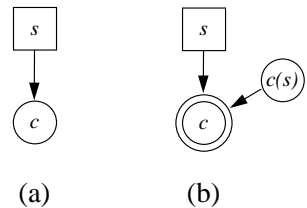

Figure 5: A transformation to Howard Canonical Form.

Table 1: The four states of the mechanism $c(s)$, which relates smoking and lung cancer.

|  | state 1 | | state 2 | | state 3 | | state 4 | |
|---|---|---|---|---|---|---|---|---|
| smoke | no | yes | no | yes | no | yes | no | yes |
| lung cancer | no | yes | yes | no | no | no | yes | yes |

Second, additional assessments typically are required in order to transform an influence diagram into HCF. For example, only one independent probability assessment is needed to quantify the influence diagram in Figure 5a, whereas three independent assessments are required for the node $c(s)$ in Figure 5b. We return to this point later in this section.

Third, we can think of the $c(s)$ as the *causal mechanism* that relates smoking and lung cancer.[5] Although we may not be able to observe this mechanism, we note that a clairvoyant always can. Consequently, this mechanism passes the clarity test.

Finally, although we may not be able to observe $c(s)$ directly, we may be able to learn something about the mechanism. For example, we can imagine a test that measures the susceptibility of someone's lung tissue to lung cancer in the presence of tobacco smoke. The probabilities on the outcomes of this test would depend on $c(s)$, and this test may be done prior to making the decision *smoke*. Indeed, with the influence diagram now in HCF, we can compute the value of information of $c(s)$ to investigate whether or not this test would be cost effective.

Let us consider another example of a transformation to HCF. In the influence diagram shown in Figure 6a, the variable *lung cancer* depends on *smoke* and *genotype*. Therefore, we could extract the uncertainty in this relationship by introducing a variable that represents lung cancer as a function of *smoke* and *genotype*. The variable *genotype*, however, is already in the fixed set with respect to decisions. Therefore, we need only extract the uncertainty associated with the relationship between *smoke* and *lung cancer*, introducing the causal mechanism $c(s)$. As in our previous example, *lung cancer* is a deterministic function of *smoke* and $c(s)$. In contrast to our previous example, however, $c(s)$ depends on *genotype*. A similar transformation can be performed on the variable *cardiovascular status* and its parents, yielding the HCF influence diagram shown in Figure 6b. In this influence diagram, all chance variables are in the fixed set; and we may compute their value of information.

---

[5]Pearl (1994) calls $c(s)$ a *disturbance*.

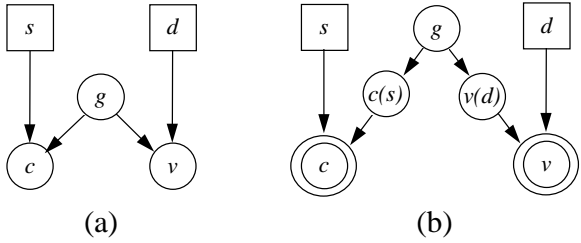

Figure 6: Another transformation to Howard Canonical Form.

To transform an arbitrary influence diagram into HCF, we want to extract the uncertainty associated with each variable into an associated causal mechanism. As we have discussed, the causal mechanism for $x$ need include only those parents of $x$ that are not in the fixed set with respect to decisions.

**Definition 9 (Mechanism)** *Given a causal influence diagram with uncertain variables $U$ and decisions $D$, a* mechanism *for $x \in U \setminus F(D)$ with respect to $D$ is a new variable $x(Y)$ which represents all possible mappings from $Y = Pa(x) \setminus F(D)$ to $x$.*

If $x$ is in the fixed set $F(D)$, then we cannot recognize a causal mechanism associated with $x$. If $x$ is not in the fixed set $F(D)$, then a subset of the variables $Y$ in the definition (usually, $Y$ itself) will be causes for $x$. Also, by this definition, $x$ will always be a deterministic function of $x(Y)$ and $Y$.

Again, we emphasize that additional probability assessments are required for the created nodes. If $x$ has $r$ states and $Y$ has $q$ instances, then $x(Y)$ will have $r^q$ states. Thus, in general, the assessment of the probabilities associated with a causal mechanism is formidable. In real-world domains, however, reasonable assertions of independence often facilitate such assessments. In some cases, no additional assessments are necessary (see, e.g., Heckerman et al. 1994).

**Theorem 5 (Howard Canonical Form)** *Any influence diagram for chance nodes $U$ and decision nodes $D$ may be transformed into Howard Canonical Form as follows:*

1. *If there is a node not in $F(D)$ pointing to a node in $F(D)$, then reassess the influence diagram using a variable ordering for $U$ where the nodes in $F(D)$ come first*

2. *Add enough arcs to make the influence diagram a causal influence diagram*

3. *For every chance node $x$ not in $F(D)$,*
    - *Add the mechanism node $x(Y)$ to the diagram*
    - *Make $x$ a deterministic function of $Y \cup x(Y)$*
    - *Make $Pa(x) \setminus Y$ parents of $x(Y)$*

4. *Assess any additional dependencies among the variables now in the fixed set $F(D)$*

**Proof:** After steps 1 (and 2), no nodes in the fixed set $F(D)$ will be descendants of decision nodes. Therefore, after step 3, all nodes that are descendants of decisions will be deterministic nodes. Also, every mechanism node added to the diagram will be in the new fixed set $F(D)$, by definition. Thus, all nodes downstream from decisions will be (possibly indirect) deterministic functions of elements of $F(D)$ and $D$. Consequently, every node $x$ will be in $F(D|Pa(x))$; and hence the influence diagram will be causal. □

The need for step 4 in the construction is illustrated by our example in Figure 6. If we had chosen not to represent the variable *genotype*, we would have added arcs—say—from *smoke* and *lung cancer* to *cardiovascular status*, as in Figure 4b. The construction described in the theorem, ignoring the last step, would have created parentless mechanisms $c(s)$ and $v(s,c,d)$. These variables, however, would be dependent and this dependency would need to be assessed.

## 6 Counterfactual Reasoning

Causal models have been used to answer counterfactual queries [Balke and Pearl, 1994]. A counterfactual query is of the form: "if $a$ were true, then what is the probability that $b$ would have been true, given that we know $c$?" In our lifestyles decision problem, a counterfactual query would be: "given that I have not smoked, have maintained a good diet, have not gotten lung cancer, and my cardiovascular status has been good, what is the probability that I would have gotten lung cancer had I smoked and eaten poorly?" As we discuss shortly, methods for counterfactual reasoning are important, because often they can be applied to real decision problems.

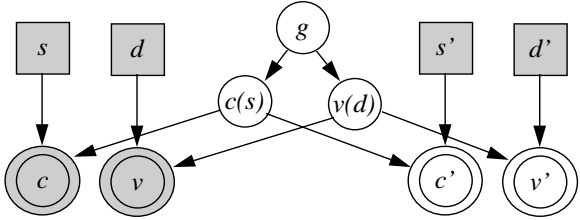

Figure 7: The use of HCF to compute a counterfactual query. The primed variables denote counterfactuals. Shaded variables are instantiated.

Influence diagrams in HCF can also be used to answer counterfactual queries. For example, to answer our lung-cancer query, we begin with the influence diagram in HCF shown in Figure 6b. Then, we make two copies of all variables not in the fixed set, as shown in Figure 7. The first copy represents the actual state of affairs; and the second copy represents the counterfactuals (in our example, $smoke = yes$ and $diet = poor$). The variables in the fixed set are not copied, because, by definition, they cannot be affected by decisions. Also, by definition of mechanism, each copy of an observable variable has the same deterministic relationship with its mechanism. To answer our query, we instantiate the decision and chance variables in the first copy of the diagram to their observed values (no smoking, good diet, no lung cancer, and good cardiovascular status, in our example). In addition, we instantiate our counterfactual decisions in the second copy of the diagram. Then, we use a standard belief-network inference method to compute the probability of the variable(s) of interest (*lung cancer* in our example).

Using this approach, we can answer arbitrary counterfactual queries, including queries where variables in the fixed set have been observed. For example, we can answer the query, "given that I have not smoked, have maintained a good diet, and have a genotype predisposing me to lung cancer, what is the probability that I would have gotten lung cancer had I smoked." The approach is closely related to that described by Balke and Pearl (1994).

Procedures for counterfactual reasoning often can be used for real decision problems. For example, consider a modification of our original counterfactual query: "Given that I have not smoked and have maintained a good diet over the last year, and have not gotten lung cancer and have a good cardiovascular status, what is the probability that I will get lung cancer one year from now if I begin to smoke?" If we assume that the relationships in the original domain model do not change over the two-year time period in the query, then we may use the influence diagram in Figure 7 to answer this query. Heckerman et al. (1994) provide another example in the domain of logic-circuit troubleshooting. Also, see the discussion in Goldszmidt and Darwiche (1994).

## 7 Global Causal Models

Most previous work on the graphical representation of causality concerns the situation where all interactions in a domain are causal (see, e.g., Pearl and Verma 1991, Druzdzel and Simon 1993, and Spirtes et al. 1993, and Pearl 1994). Here, we consider this special case, and describe correspondences between our work and the work of Pearl et al., which is representative of this body of work.

When all interactions among variables in an influence diagram are causal, we call that influence diagram a causal network.

**Definition 10 (Causal Network)** *An influence diagram with uncertain variables $U$ and decisions $D$ is said to be a* causal network *for $U$ with respect to $D$ if $Pa(x)$ are causes for $x$ with respect to $D$ for all $x \in U$.*

It follows immediately that every causal network is a causal influence diagram.

An example of a causal network is the influence diagram in Figures 2b. (Of all the influence diagrams presented in this paper, only this one is a causal network.) As another example, we can transform the influence diagram of Figure 1 into a causal network by adding the decision node *retroviral therapy* and an arc from this node to *genotype*.

As in the local case, we cannot always identify causal networks using our graphical blocking condition. From Theorem 4, however, we can do so given enough set decisions. In the following corollary, a node with at least one parent but without any children is called a *leaf* node.

**Corollary 6** *A minimal causal influence diagram for uncertain variables $U$ and decisions $D$ such that $D$ includes a set decision for each nonleaf uncertain variable in $U$ is a causal network.*

Pearl et al. define a causal model (or causal network) for a set of uncertain variables $U$ to be a minimal belief network such that every variable is caused by its parents. They take cause and effect to be a primitive notion, and do not define it. They assert that, given a causal model for $U$, it is likely that there exists a corresponding set decision for every variable in $U$.

Our approach is the reverse of theirs. We start with a definition of causality in terms of decisions, and then show that given set decisions for all nonleaf variables, all interactions must be causal.

Another correspondence exists. Namely, if we transform a causal network by our definition into HCF, we obtain a model where every chance node is a deterministic function of its old parents and a single mechanism. If we assume these mechanisms are independent, we obtain Pearl et al.'s definition of a *causal theory* for $U$.

## 8   Future Work and Conclusions

An important aspect of causality that we have barely touched upon in this paper is the notion of time. Although many of the results presented here are applicable to time-varying domains, where two different nodes in an influence diagram may represent the same system variable at different points in time, there are aspects of such domains that we have yet to explore.

We have presented a practical definition of cause and effect in precise decision-analytic terms. We have shown how the influence diagram is sometimes inadequate for the graphical representation of cause (by our definition), and have shown how some inadequacies can be eliminated. We hope that this work will begin to knit the closely related threads of research in decision analysis and causal modeling.

### Acknowledgments

We thank Jack Breese, Eric Horvitz, Ron Howard, Mark Peot, Glenn Shafer, and Patrick Suppes for useful discussions.